# Title: "The Era of Foundation Models in Medical Imaging is Approaching": A Scoping Review of the Clinical Value of Large-Scale Generative AI Applications in Radiology

Inwoo Seo[1], Eunkyoung Bae[2], Joo-Young Jeon[3], Young-Sang Yoon[4], Jiho Cha[5]

## Abstract

Social problems stemming from the shortage of radiologists are intensifying, and artificial intelligence is being highlighted as a potential solution. Recently emerging large-scale generative AI has expanded from large language models (LLMs) to multi-modal models, showing potential to revolutionize the entire process of medical imaging. However, comprehensive reviews on their development status and future challenges are currently lacking. This scoping review systematically organizes existing literature on the clinical value of large-scale generative AI applications by following PCC guidelines. A systematic search was conducted across four databases: PubMed, EMbase, IEEE-Xplore, and Google Scholar, and 15 studies meeting the inclusion/exclusion criteria set by the researchers were reviewed. Most of these studies focused on improving the efficiency of report generation in specific parts of the interpretation process or on translating reports to aid patient understanding, with the latest studies extending to AI applications performing direct interpretations. All studies were quantitatively evaluated by clinicians, with most utilizing LLMs and only three employing multi-modal models. Both LLMs and multi-modal models showed excellent results in specific areas, but none yet outperformed radiologists in diagnostic performance. Most studies utilized GPT, with few using models specialized for the medical imaging domain. This study provides insights into the current state and limitations of large-scale generative AI-based applications in the medical imaging field, offering foundational data and suggesting that the era of medical imaging foundation models is on the horizon, which may fundamentally transform clinical practice in the near future.

[1] Korea Advanced Institute of Science and Technology(KAIST), Graduate school of Future strategy, Room N5-2259, 291, Daehak-ro, Yuseong-gu, Daejeon, Republic of Korea 34141

[2] Korea Advanced Institute of Science and Technology(KAIST), Graduate school of Future strategy, Room N5-2259, 291, Daehak-ro, Yuseong-gu, Daejeon, Republic of Korea 34141

[3] Korea Advanced Institute of Science and Technology(KAIST), Graduate school of Future strategy, Room N5-2259, 291, Daehak-ro, Yuseong-gu, Daejeon, Republic of Korea 34141

[4] Korea Advanced Institute of Science and Technology(KAIST), Graduate school of Future strategy, Room N5-2259, 291, Daehak-ro, Yuseong-gu, Daejeon, Republic of Korea 34141

[5] Korea Advanced Institute of Science and Technology(KAIST), Graduate school of Future strategy, Room N5-2259, 291, Daehak-ro, Yuseong-gu, Daejeon, Republic of Korea 34141, E-mail address: chajiho@kaist.ac.kr

# 1. Introduction

The persistent shortage of radiologists, exacerbated by increasing demand for medical imaging, poses significant challenges. The number of radiologists has not kept pace with this growing demand, and many specialists are retiring without enough new ones being trained, leading to increased misdiagnosis, unnecessary medical tests, and higher healthcare costs.[1] This issue is becoming more acute due to the aging population, overworked radiologists, the rise of 3D medical imaging technologies like CT and MRI, and the increasing number of people with health insurance. Proposed solutions include the utilization of artificial intelligence (AI)[2] remote reading diagnostic technologies, and easing immigration barriers for foreign specialists.

AI diagnostic solutions based on deep learning, particularly using Convolutional Neural Network (CNN) architectures, are promising in medical imaging [3]. CNNs are effective at image processing by analyzing pixels to understand the overall image [4,5]. However, CNNs have limitations [6]: they excel in learning local patterns but struggle with long-term dependencies crucial in medical imaging. Other issues include labor-intensive research and development requiring annotation and difficulties in understanding the global context of an image based on a fixed receptive field, impacting diagnostic accuracy, especially for large or multiple lesions [7,8].

In 2017, the Transformer architecture revolutionized AI technology with the self-attention mechanism introduced in the paper "Attention Is All You Need".[9] This advancement significantly improved natural language processing performance, underpinning models like GPT and BERT and enhancing AI accessibility and application. The October 2022 launch of OpenAI's ChatGPT showcased these advancements, making AI more accessible and interactive, with continuous improvements through Reinforcement Learning Human Feedback (RLHF)[10]. In March 2023, OpenAI released GPT-4, a multimodal model capable of processing text and image input, expanding applications in education, healthcare and entertainment(etc.). Subsequently, OpenAI introduced GPT-4V, DALL-E3, CLIP, Whisper, SORA, and GPT-4o, demonstrating AI's ability to process diverse data types. Healthcare-specific AI models like Google's Med-PaLM also emerged, with Med-PaLM scoring over 60% on USMLE-style questions[11] and Med-PaLM 2 scoring 85%[12], integrating various medical data to enhance patient care. Recently, Med-Gemini further established its role in healthcare.[13]

Despite these advancements, there are no systematic studies on clinical value [14] of the application of LLM or multimodal generative AI technologies in the field of radiology.[15] The rapid developments in AI have the potential to transform medical imaging, prompting several key questions:

- What is the clinical value of AI applications in medical imaging?
- How are these technologies being categorized and evaluated by clinicians?
- What opportunities and challenges exist for AI in this field?

# Result

Our database and hand search identified a total of 14,370 articles, with additional records from ArXiv (n=1), ACM (n=1), and MDPI (n=1). After and initial screening based on primary & secondary criterions, 73 articles considered eligible for full-text screening. (see Fig. 1 for details). The final sample of peer-reviewed articles entering the analysis included a total of 15 studies, described in Table 1. A list of references for the included studies is available in Supplementary Note 1.

**Figure 1: Literature selection process**

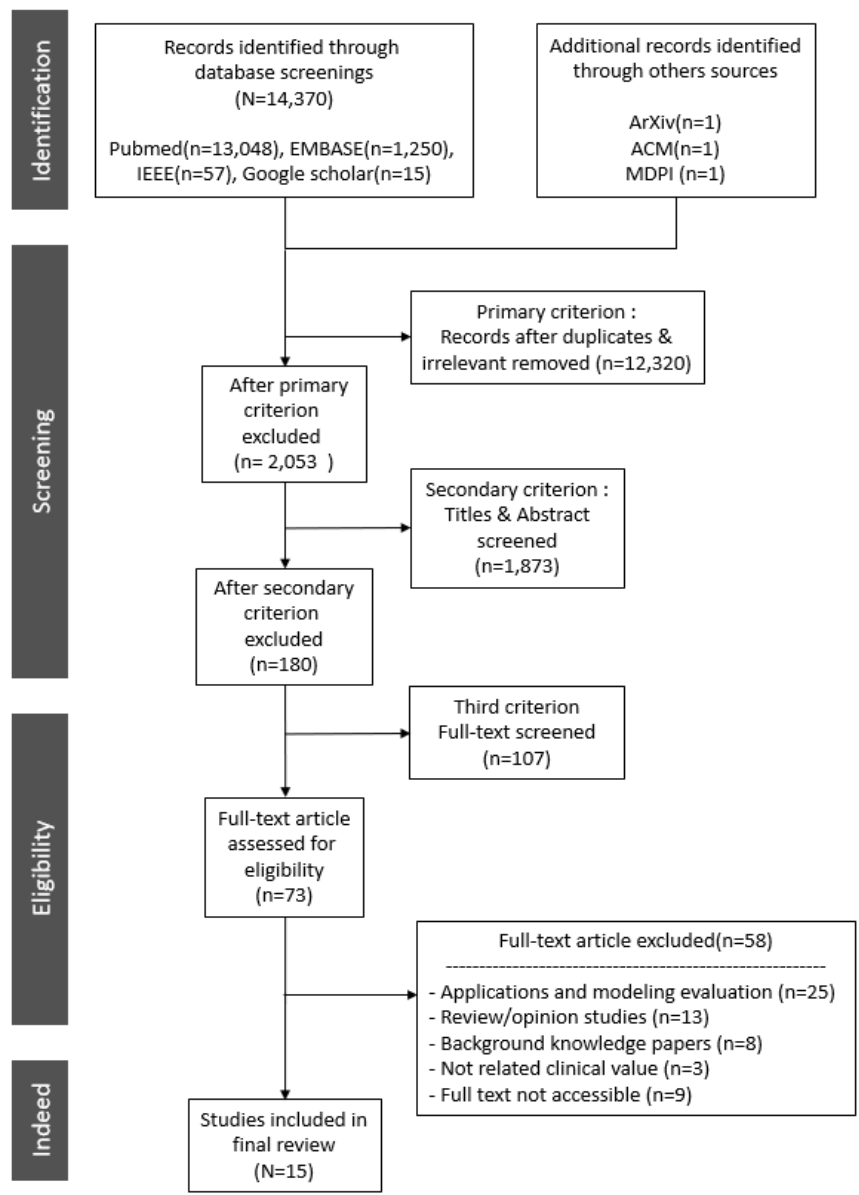

## Study characteristics

We searched for studies from 2017 through April 2024, when the transformer architecture first began to emerge. In terms of researchers, radiologists authored the most targeted articles (13), followed by oncologists and Ph.D researchers belongs to radiology department with one article each. In terms of publication year, 9 articles were published in 2023 and 6 in 2024, reflecting the latest research trends, and 7 articles were published in specialized radiology journals (Radiology, JCR, ACR, ECR), 4 in medical-related journals (Cureus, Sage Journal, MDPI), 2 in technical journals (IEEE, ACM), and 2 in other journals (springer open, arXiv). In terms of the purpose of the study, 10 studies were based on 'Radiology report', which accounted for the largest proportion, followed by 2 studies related to 'medical imaging findings and patient medical information', 2 studies related to 'medical imaging images', and 1 study related to 'radiologists'. In terms of both domestic and overseas, North America (the United States) accounted for the largest number of papers with 7, followed by Asia (India, Japan, China, Korea, and Australia) with 5, and Europe (Germany and the United Kingdom) with 3.

**Table 1: General characteristics of the literature studied**

| No | Author | Researchers | Year | Country | Journal name | Research objectives | Types of studies | Research Materials |
|---|---|---|---|---|---|---|---|---|
| 1 | Gertz RJ et al.[16] | Radiologists | 2024 | Germany | Radiology | Evaluate the effectiveness of GPT-4 in identifying medical image readout errors from a performance, time, and cost-effectiveness perspective | Retrospective studies | Medical Imaging Readings |
| 2 | Takeshi Nakaura et al.[17] | Radiologists | 2024 | Japan | Japanese Journal of Radiology | Evaluate the ability of GPT series to generate medical image readings and compare performance to reports written by radiologists | Retrospective studies | Medical Imaging Readings |
| 3 | Reuben A. Schmidt et al.[18] | Radiologists | 2024 | Austrailia | Radiology | Evaluate the performance of five major generative LLMs (GPT-3.5-turbo, GPT-4, text-davinci-003, Llama-v2-70B-chat, Bard) in automatically detecting speech recognition errors in medical image transcripts | Retrospective studies | Medical Imaging Readings |
| 4 | Zhaoyi Sun, MS et al.[19] | Radiologists | 2023 | United States | Radiology | Evaluating Medical Imaging Impression Generation with GPT-4 | N/A | Medical Imaging Readings |

| No | Author | Researchers | Year | Country | Journal name | Research objectives | Types of studies | Research Materials |
|---|---|---|---|---|---|---|---|---|
| 5 | Emile B. Gordon MD et al.[20] | Radiologists | 2024 | United States | American College of Radiology | Evaluate accuracy, relevance, and readability of ChatGPT's patient imaging questions and validate the effectiveness of simple prompts | N/A | Other than medical imaging findings |
| 6 | Pokhraj P. Suthar et al.[21] | Radiologists | 2023 | United States | Cureus | Evaluation of ChatGPT 4.0's diagnostic accuracy when patient history and imaging findings are provided | N/A | Other than medical imaging findings |
| 7 | Katharina Jeblick et al.[22] | Radiologists | 2023 | Germany | European Radiology | Evaluate the quality of summarized medical image readings generated using ChatGPT | Prospective studies | Medical Imaging Readings |
| 8 | Qing Lyu et al.[23] | Radiologists | 2023 | United States | Springer Open | Evaluation of ChatGPT's performance in translating medical image readings into layman (patient) language | N/A | Medical Imaging Readings |
| 9 | Eric M Chung et al.[24] | Oncologists | 2023 | United States | Sage Journal | Evaluate ChatGPT's ability to generate MRI summary reports for prostate cancer patients and survey physician satisfaction with AI summary reports | N/A | Medical Imaging Readings |
| 10 | Huan Jiang et al.[25] | Radiologists | 2024 | China | European Journal of Radiology | Evaluation of accuracy and reproducibility of structured thyroid ultrasound reading generation using ChatGPT | Retrospective studies | Medical Imaging Readings |
| 11 | Pradosh Kumar Sarangi et al.[26] | Radiologists | 2023 | India | Cureus | Evaluate ChatGPT-3.5's ability to simplify medical image readings | N/A | Medical Imaging Readings |
| 12 | Batuhan Gundogdu et al.[27] | Department of Radiology Affiliations Researcher (Ph.D) | 2023 | United States | IEEE | Research on a deep learning approach to automate the generation of impressions from the analysis of medical imaging findings and patient information. | N/A | Medical Imaging Readings |
| 13 | Nur Yildirim et al.[28] | Radiologists | 2024 | United Kingdom | ACM | Clinical usability evaluation and design considerations for visual language model (VLM) interactions with medical images - with a focus on radiologists' evaluations | N/A | Radiology Specialty |

| No | Author | Researchers | Year | Country | Journal name | Research objectives | Types of studies | Research Materials |
|---|---|---|---|---|---|---|---|---|
| 14 | Ro Woon Lee et al.[29] | Radiologists | 2023 | South Korea | MDPI | Evaluating diagnostic accuracy and clinical utility of chest X-ray reading using large-scale artificial intelligence and large language models | N/A | Medical Video Image |
| 15 | Zhengliang Liu et al.[30] | Radiologists | 2023 | United States | arXiv | Introducing a large-scale study evaluating the capabilities and limitations of GPT-4V for biomedical image analysis. | N/A | Medical Video Image |

## Technical Characteristics of AI technologies

We analyzed the technical characteristics and types of language models (LLMs) and multimodal models used in the studies.<table 2> In terms of the types of A.I. models, language models (LLMs) accounted for the largest share with 12 studies, while multimodal visual language models (VLMs), a relatively new technology, accounted for 3 studies. In terms of AI developers and models, the leading company, OpenAI's GPT, dominated with 12 studies, followed by Google, Meta, open source public models (BERT, LSTMs), and Kakao Brain's KARA-CXR model with one study each. In terms of medical imaging modality types, 3D modalities such as CT/MRI accounted for the majority with 9 articles, followed by chest X-ray with 4 articles, ultrasound with 2 articles, and others.

**Table 2: Technical characteristics and types of Generative AI**

| No | Types of Models | AI Developer | Modality | AI Model Versions |
|---|---|---|---|---|
| 1 | Language model (LLM) | OpenAI | CT/MRI | GPT-4.0 |
| 2 | Language model (LLM) | OpenAI | CT | GPT-2.0, GPT-3.5, and GPT-4.0 |
| 3 | Language model (LLM) | OpenAI and more | CT/MRI | GPT-3.5-turbo, GPT-4, Text-davinci-003, Llama-v2-70B-chat, and Bard |
| 4 | Language model (LLM) | OpenAI | Chest radiograph | GPT-4.0 |
| 5 | Language model (LLM) | OpenAI | N/A | GPT-3.5 |
| 6 | Language model (LLM) | OpenAI | CT | GPT-4.0 |
| 7 | Language model (LLM) | OpenAI | CT/MRI | GPT-4.0 |
| 8 | Language model (LLM) | OpenAI | CT | ChatGPT, GPT-4.0 |
| 9 | Language model (LLM) | OpenAI | MRI | GPT-3.5 |
| 10 | Language model (LLM) | OpenAI | Ultrasound | gpt-3.5, gpt-4.0 |
| 11 | Language model (LLM) | OpenAI | CT/MRI | GPT-3.5 |
| 12 | Language model (LLM) | Google | CT/MRI | BERT |
| 13 | Visual Language Model (Multimodal) | Microsoft | Chest radiograph | VLMs: Multimodal Foundation Models |
| 14 | Visual Language Model (Multimodal) | Open AI, Kakao Brain | Chest radiograph | kara-cxr, gpt-4v |
| 15 | Visual Language Model(Multimodal) | OpenAI | Chest radiograph and others | GPT-4V |

## Clinical Value of generative AI application in Radiology

Table 3. shows the clinical value of the research object and the results of the research in the literature. The clinical value was categorized into three categories: Accuracy[31,32,33] Workflow Efficiency[34], and Better Outcomes for Patients.[35]

1. Accuracy

Accuracy was described in a total of eight papers, with details on detecting various types of errors in the readings and improving diagnostic performance. Regarding error detection, a study (No. 1) was conducted to determine how much the detection of errors caused by low-experience doctors, error detection due to limitations in voice recognition, and mistakes due to excessive workload could be improved by utilizing the LLM algorithm, and as a result, the detection rate of GPT-4 (82.7%) was similar to that of radiologists. In a study that focused solely on errors due to voice recognition in the reading text, GPT-4's detection performance was confirmed to be highly accurate, with F1 scores of 94.3% and 86.9%, respectively, after classifying clinically significant and insignificant voice recognition errors.

In terms of diagnostic performance, studies have been conducted to compare the reading capabilities of LLMs and radiologists under certain conditions and between different LLMs, and the results show that radiologists' capabilities are generally higher than those of AI models when it comes to patient information-based reading performance.(No.2)) The performance of generating impressions in the reading was also compared with AI, and it was concluded that radiologists were still better in terms of clarity and accuracy.(No.4) However, in an experiment to measure the improvement of AI's diagnostic performance while providing additional patient information, the performance gradually improved as more information was added. (Accuracy increased by 10% after 2 weeks, 23.57% after 3 weeks, and 57.86% after 4 weeks) (No.6). In a study on diagnostic performance between the multimodal vision language model GPT-4V and KARA-CXR (No.14), KARA-CXR was found to be about 70% accurate in reading, outperforming ChatGPT's 45%. In addition, the false positive rate was 68% for KARA-CXR and 37% for ChatGPT, and the non-hallucination rate (wrong prediction) was 75% for KARA-CXR and 38% for ChatGPT, showing KARA-CXR's overall superior performance.

2. Streamline workflow

Six studies were conducted on workflow efficiency, mainly related to time, cost, and improving the quality and ease of reading. More specifically, one study (NO.1) evaluated the diagnostic processing speed and cost of GPT-4 compared to radiologists, and found that GPT-4's processing speed was 3.5 seconds (radiologists: 25.1 seconds) and the cost of correction was $0.03 (radiologists: $0.42), which can enhance the cost-effectiveness of the workflow. A study on radiologists' satisfaction after generating a conversion from a free text report to a structured report between AI models (GPT-4 and GPT-3.5) was also conducted (No. 10).), GPT-3.5 generated a total of 202/236 (74.3%) satisfactory structured reports, while GPT-4.0 generated a total of 69/236 (25.4%) satisfactory structured reports, indicating that GPT-3.5 performed better in terms of the quality of the readings.

Another perspective study (No. 12) asked radiologists to rate each of 1,000 AI-generated impressions and a human-generated report on the generation of in-reading impressions, with a "negative" rating if the original was better, a "positive" rating if the AI-generated impression was better, and a "neutral" rating if they were about the same, The results showed that 669 reports were labeled "neutral," 83 were labeled "positive," and 235 were labeled "negative," with 76% of the predictions being as accurate as the human-generated findings. There were also three studies (Nos. 13, 14, and 15) on vision language models that showed that they could improve workflows, with one (No. 14) in particular.) compared the accuracy and usability of two AI systems for reading chest X-rays, KARA-CXR and ChatGPT, and evaluated a total of 2,000 cases, and found that KARA-CXR was more accurate, about 70%, while ChatGPT was about 45%, In terms of false positives, KARA-CXR had fewer false positives, approximately 68% versus 37% for ChatGPT, and in terms of false positives, KARA-CXR had a 75% non-hallucination rate versus 38% for ChatGPT, leading to the overall conclusion that KARA-CXR was superior. Another study (No. 13) evaluated 13 radiologists on the evaluation of the chest X-ray multimodal vision language foundation model alone and provided four main qualitative feedback to improve workflow. First, draft reading generation would only be meaningful if the accuracy was near perfect, with evaluators saying they would only be willing to revise one or fewer reports out of every 10 draft readings generated by the AI. Second, regarding workload and time, evaluators noted that reading and reporting on multislice images such as CT is time-consuming, so AI draft readings have the potential to save time and reduce cognitive load. Third, when it came to report format, short, standardized readings were preferred, with all radiologists reporting a preference for bullet point-style reports (Structured reports) over free-text reports. Fourth, regarding the prioritization of reading list findings and the presentation of reliability (quantified scores), they preferred to prioritize findings according to clinical importance and to present reliability as a score out of 5. (NO. 15) In the case of the study, GPT-4V was analyzed with datasets within 16 different imaging modalities (X-ray, CT, MRI, etc.), and representatively, the study related to chest X-ray used public datasets (MIMIC, CheXpert), and after specifying 13 representative lesions with specific prompts and images, the results of the experiments on various tasks were as follows. Multi-class classification failed to correctly recognize most diseases, with low accuracy for most diseases except Cardiomegaly, and binary classification for certain diseases, such as COVID-19, performed better than multi-class classification, but still showed significant errors. Disease localization tasks struggled to accurately recognize small lesion areas, while large lesion areas such as COVID-19 were approximated. In terms of diagnostic report generation, GPT-4V performed best in image-based report generation, showing a high degree of semantic alignment with the actual report.

3. Patient usability

A total of five studies were conducted on Better outcomes for patients, mostly related to improving doctor-patient communication. The most common topic covered was radiologists' ratings of the factual correctness, consistency, completeness, readability, and potential harm of large language models (LLMs) that translate and summarize medical terms into language that patients can understand. The results of the studies were generally divided into positive and negative evaluations, with a total of three studies (No 7, No 8, and No 11) reporting on positive evaluations and two studies (No 5 and No 9) reporting on negative evaluations.

First, a study (No. 7) evaluating the quality of simplified readings generated by the large language model (LLM) ChatGPT was positive about ChatGPT's performance in summarizing readings, with more than 75% of readings "agreeing" or "strongly agreeing" with the statement that the report was factually accurate and complete across all quality criteria. Meanwhile, a study using ChatGPT/GPT to translate readings into language that is easier for patients and providers to understand (No.8), ChatGPT's translation score was 4.268 out of 5, and it was found that unambiguous and optimized prompts increased the performance for key information by 77.2%. In addition, in a study (N0.11) on the performance of GPT-3.5 in translating readings to a level that is easy for patients to understand, GPT-3.5 achieved the highest accuracy (94.17%) in detailing readings and the lowest accuracy (85%) in drawing patient conclusions, which was also positively evaluated for usability. However, the study also found that ChatGPT was not suitable for patient communication when translated from Hindi to English. On the other hand, a study (No. 5) on whether the summaries provided by ChatGPT improve performance with or without prompts concluded that while prompts reduced variability in responses and provided more targeted information, they did not improve readability, and none of the summaries reached the recommended $8^{th}$ grade reading level for patient comprehension.

Another study (No. 9) found that ChatGPT-summarized reports were more readable. The majority of respondents (oncologists) rated the summarized reports as factually accurate (89%), easy to understand (78%), and complete (93%). However, they were divided on the likelihood of the report causing harm to the patient (51%), overall quality (53%), and which physician to send it to (46%). In conclusion, the results suggest that while respondents (oncologists) rated AI summarized reports highly for accuracy, comprehensibility, and completeness, they are not yet fully satisfied with them in terms of potential harm, quality, and patient deliverability.

**Table 3: Clinical value and study results**

| No | Clinical Value | | | Research Findings |
|---|---|---|---|---|
| | Accuracy | Workflow Efficiency | Better Patient Outcome | |
| 1 | ○ <br> Error Detection | ○ <br> Time, money | | - Evaluate the effectiveness of GPT-4 in identifying medical image readout errors from a performance, time, and cost-effectiveness perspective <br> - Detection rate for GPT-4 (82.7%): Similar to radiologists (senior: 89.3%; fellows and residents: 80.0%); <br> - Only one senior radiologist performed better.(94.7%). <br> - Processing speed for GPT-4: 3.5 seconds/report (radiologists: 25.1 seconds/report), Revision cost: $0.03/report (radiologists: $0.42/report). |
| 2 | ○ <br> Diagnostic Performance | | | - Accuracy evaluation between specialists and GPTs: Specialists ranked highest in Top-1 and Top-5, accuracy, followed by GPT-4, GPT-3.5, and GPT-2. <br> - No significant differences in grammar, readability, image findings, or overall quality between radiologists and GPT-3.5 and GPT-4. <br> - However, GPT models score lower on Impression and Differential Diagnosis |
| 3 | ○ <br> Error Detection | | | - GPT-4 detects clinically important speech recognition errors (F1 score, 86.9%) and non-critical errors (F1 score, 94.3%) with high accuracy. <br> - GPT-4 effectively flagged serious errors such as internal inconsistencies (12/12, 100%) and nonsensical wording (22/24,91.7%) that require fluency in the context of the entire report and radiology text. <br> - The increased error rate is due to the fact that for longer reports, practitioner speech recognition (Dictation) and Overnight Shift are associated with the and identified as highly productive as an error correction tool in this area. |
| 4 | ○ <br> Diagnostic Performance | | | - Provide Findings Evaluate comparisons for generating Impressions <br> - Radiologist-generated findings are rated higher than GPT-4-generated findings for quality and safety due to their clarity and accuracy. <br> - However, referring clinicians rated the findings generated by GPT-4 as more consistent and less harmful than radiologists thought. |

| No | Clinical Value | | | Research Findings |
|---|---|---|---|---|
| | Accuracy | Workflow Efficiency | Better Patient Outcome | |
| 5 | | | ○ | - Analyzed patient responses to common imaging-related questions to assess the accuracy, relevance, and readability of ChatGPT and to investigate the impact of simple prompts. |
| | | | Increase patient communication | - Of the 264 responses, the accuracy rate was 83% for those without prompts and 87% for those with prompts (P = .2).<br>- Consistency improved from 72% to 86% due to the use of prompts (P =.02).<br>- Almost all responses (99%) were at least partially relevant, with fully relevant responses increasing from 67% to 80% due to the use of prompts (P = .001).<br>- The average readability level was high at 13.6 grade levels, which did not change with the use of prompts, and no responses met the recommended level of 8$^{th}$ grade for patients. |
| 6 | ○ | | | - ChatGPT's diagnostic accuracy depends on the amount of information provided. |
| | Diagnostic Performance | | | - Patient history alone was 4.29% accurate, the same when imaging findings from the first week were included.<br>- Accuracy increases to 10% when including 2 weeks of imaging findings.<br>- Accuracy increases to 23.57% when including 3 weeks of imaging findings.<br>- When including 4 weeks of medical history and imaging data, the accuracy rate is 57.86%.<br>Increase |
| 7 | | | ○ | - Evaluate the quality of simplified readings generated with large language model (LLM) ChatGPT |
| | | | Increase patient communication | - Fifteen radiologists with 5 years of median experience (IQR 1 10 years) rated this simplified reading on a 5-point Likert scale and provided free-form comments.<br>- Evaluation results for 45 simplified reports (Table 2; Fig. 3a): Generally agreed with the statement that the reports were factually accurate and complete (median = 2) More than 75% "agreed" or "strongly agreed" on both quality criteria (Q3 = 2) "Strongly disagreed" was not selected at all Disagreed that incorrect conclusions from simplified reports were likely to result in physical and/or psychological harm (median = 4) No radiologists selected "strongly agreed" |

| No | Clinical Value | | | Research Findings |
|---|---|---|---|---|
| | Accuracy | Workflow Efficiency | Better Patient Outcome | |
| 8 | | | ○ | - Investigating the feasibility of experimenting with ChatGPT to translate readings into language that is easier for patients and providers to understand<br>- ChatGPT's translation score: 4.268 out of 5. |
| | | | Increase patient communication | - Average missing information points per translation: 0.097, misinformation points: 0.065.<br>- However, ChatGPT tends to oversimplify or omit key points.<br>- Only 55.2% of key points were fully translated due to ambiguous prompts.<br>- Increases to 77.2% with optimized prompts.<br>- ChatGPT vs. GPT-4: GPT-4 significantly improves translation quality |
| 9 | | | ○ | - The study evaluated the readability and quality of the ChatGPT-generated MRI report summary compared to the full MRI report.<br>- Median readability score (FK score) for the full report was 9.6 Vs 5.0 for the executive summary ($p < 0.05$) (lower scores indicate better readability) |
| | | | Increase patient communication | - 12 radiation oncologists reviewed the summary and gave it a score (out of 5):<br>  - Factual accuracy: 4.0  - Comprehension: 4.0<br>  - Completeness: 4.1     - Potential harms: 3.5<br>  - Overall quality: 3.4<br>  - Likelihood to communicate to patients: 3.1<br>- Physicians are more likely to be satisfied with summarized reports in terms of factual accuracy, ease of understanding, and completeness.<br>  However, they are less likely to be satisfactory in terms of likelihood of harm to the patient, overall quality, and likelihood of being sent to the patient. |
| 10 | ○ | ○ | | - The study compared the performance of GPT-3.5 and GPT-4.0 in generating structured reports from 136 free text ultrasound reports.<br>- GPT-3.5 Performance: 202 satisfactory structured reports (74.3%); GPT-4.0 Performance: 69 satisfactory structured reports (25.4%) |
| | Diagnostics Performance | Improve reading quality and authoring ease | | - Thyroid nodule classification accuracy: GPT-4.0: 69.3%, GPT-3.5: 34.5%<br>- Key takeaway: ChatGPT-4.0 is better at classifying and managing thyroid nodules. ChatGPT-3.5 had higher accuracy and consistency in recommendations, but ChatGPT-3.5 outperformed in structured report generation. |

| No | Clinical Value | | | Research Findings |
|---|---|---|---|---|
| | Accuracy | Workflow Efficiency | Better Patient Outcome | |
| 11 | | | ○ | - A study of GPT-3.5's performance in translating readings to a level that is easy for patients to understand |
| | | | Increase patient communication | - GPT-3.5 had the highest accuracy (94.17%) in describing reading details. Lowest accuracy (85%) for drawing patient conclusions. Case-specific scores are similar (p-value = 0.97).ChatGPT's Hindi translation is not suitable for patient communication. |
| | | | | - The current free version of ChatGPT-3.5 can be a useful tool for non-healthcare professionals and patients to understand readings, make clinical decisions, or answer report-related questions. |
| | | | | - The model's ability to bridge the gap between jargon and accessible communication has great potential to enhance interdisciplinary collaboration and improve patient care in a variety of healthcare settings. |
| 12 | | ○ | | - Evaluate clinical validity of predicted impressions using radiologist assessment with ROUGE analysis for statistical validation |
| | | Improve reading quality and authoring ease | | - AUC Medicine radiologist and an external, independently certified radiologist evaluated a random sample of 1,000 reports to compare original and AI-predicted findings. |
| | | | | - "Negative" if the original is better, "Positive" if the AI-generated impression is better, and "Neutral" if they are about the same. |
| | | Time, money | | - The results show that 669 reports are labeled "neutral," 83 are labeled "positive," and 235 are labeled "negative." |
| | | | | - Overall, 76% of predictions were rated as accurate as human-generated opinions, demonstrating the power of AI. |
| 13 | | ○ | | - This thesis investigates the potential utility and design requirements of utilizing vision-language models (VLMs) in radiology. |
| | | Improve reading quality and authoring ease | | - Phase 3 research included brainstorming with clinical experts and sketching four specific VLM use cases. |
| | | | | - Feedback from 13 clinicians and radiologists provided insights into clinician acceptability, desirability, and detailed design considerations |
| | | | | - Studies emphasize integrating AI into specific practical tasks to improve workflow integration and create useful mental models |
| | | | | - Research highlights tradeoffs, including balancing the usefulness of AI with the cost of risk, human effort, changes in work practices, delays in AI output, and computational requirements. |

| No | Clinical Value | | | Research Findings |
|---|---|---|---|---|
| | Accuracy | Workflow Efficiency | Better Patient Outcome | |
| 14 | ○ Diagnostic Performance Error Detection | ○ Improve reading quality and authoring ease | | - This study compares two AI systems for reading chest X-rays, Comparing the accuracy and usability of KARA-CXR and ChatGPT.<br>- Accuracy: KARA-CXR was more accurate, scoring around 70%, ChatGPT scored about 45%.<br>- False positive findings: KARA-CXR has fewer false positives, scoring around 68% received, ChatGPT scored 37%.<br>- Hallucinations (false predictions): KARA-CXR scored a 75% non-hallucination rate. ChatGPT, on the other hand, has 38%.<br>- Two radiologists read 2,000 X-rays in the evaluated, and there was good agreement on the results for both algorithms. |
| 15 | ○ Diagnostic Performance | ○ Improve readability and ease of creation | | - In this study, we evaluated the performance of GPT-4V, a state-of-the-art multimodal language model, on biomedical image analysis. 16 different medical data sets (including chest x-rays)<br>- Evaluate GPT-4V's performance through clinical challenges (image analysis, anatomical recognition, disease diagnosis, report generation, disease localization)<br>- As a result, the GPT-4V demonstrated excellent ability to distinguish between different medical imaging modalities and anatomical structures, and in some tests showed the ability to analyze biomedical research results consistent with expert knowledge.<br>- However, it has limitations when it comes to diagnosing diseases and generating comprehensive medical reports. In particular, in the worst cases, it "hallucinated" facts and overlooked important information in the image input. |

## Opportunities and Challenges of Research Subjects

The study identified several opportunity factors. AI is easy to utilize for medical image reading and error detection, demonstrating excellent performance in terms of time, cost, and accuracy. It shows promise in automatic impression generation, structured reporting, and report generation. Additionally, AI can be used to interact with patients by translating readings into understandable language, communicating medical knowledge, and overcoming communication barriers with clinicians. AI also serves as a valuable training and feedback tool for junior trainees, outperforming them in certain tasks. There is potential for developing specialized medical imaging LLM models that reflect domain characteristics, which could enhance performance further.

However, several challenges were noted. Privacy issues arise due to the cloud-based nature of LLMs, leading to concerns about medical data leakage and difficulties integrating AI systems into hospital infrastructure. Performance and accuracy gaps compared to experienced radiologists, inconsistencies in AI analysis, information oversimplification, and language performance disparities are significant issues. Finally, technical and institutional limitations, such as hallucinations, training data biases, and the inability to evolve the model through real-time learning, present ongoing challenges.<Table 4>

**Table 4: Opportunities and Challenges in the Study area**

| Separation | Opportunities | Challenges |
|---|---|---|
| 1 | - LLMs' error detection in medical imaging reports has the potential to save time and money by aligning with radiologists' opinions<br>- An offline or local server-based version of GPT-4 with API can be used as a training tool to educate residents, highlight errors, provide real-time feedback and correct medical imaging reports | - Requires domain-specific fine-tuning or workflow integration<br>- Legal and data privacy concerns are not fully addressed<br>- due to performance gaps compared to a complete medical imaging report. Human supervision is still required |
| 2 | - GPT-3.5 and GPT-4 have the potential to generate medical image readings with high readability and reasonable image findings from very short keywords<br>- GPT-4 has the advantage of being able to search the internet for the latest information, potentially giving it an advantage over other models. In addition, the various GPT models are improving | - Concerns about the accuracy of findings and differential diagnoses persist, requiring validation by a radiologist |
| 3 | - LLMs, especially GPT-4, are good at automatically detecting speech recognition errors in medical image transcripts | - These tools (LLMs) still require manual visual review of all medical imaging readings before a radiologist signs off on them<br>- Rapidly outdated due to rapidly evolving model capabilities |
| 4 | - GPT-4 and other generative AI software have the potential to revolutionize medical imaging by streamlining the production of medical image readings, which can lead to greater healthcare efficiency. | - LLMs improve consistency between findings and extrapolation sections, double-checking findings generated by radiologists |
| 5 | - ChatGPT simplifies time-consuming tasks in patient health education | - Performance of ChatGPT in languages other than English is uncertain<br>- Some knowledge areas may be underrepresented, resulting in less accurate responses<br>- Need to address readability issues and mitigate the risk of providing misleading information |
| 6 | - LLMs like ChatGPT can improve medical image interpretation and patient interaction<br>- This can help increase accuracy and speed<br>- ChatGPT fills this gap by streamlining diagnostic procedures efficiently, contributing to a more efficient healthcare delivery system | - LLMs face bias, privacy, and ethical issues<br>- Sporadic errors and "hallucinations" can occur<br>- Biased data can skew diagnoses and marginalize certain groups<br>- Patient data is essential, but privacy concerns are raised |
| 7 | - ChatGPT-like LLMs offer huge opportunity for medical text simplification, surpassing traditional machine learning approaches | - Hallucinations are inherent to LLM and difficult to address |

| Separation | Opportunities | Challenges |
|---|---|---|
|  | - Envision a future where certified ChatGPT-like LLMs are integrated into clinics and medical imaging centers | - Training data is static and not reflective of new research<br>- ChatGPT suffers from inherent bias and unbalanced data biased<br>- Non-deterministic output hinders reproducibility<br>- Privacy risks when uploading information |
| 8 | - ChatGPT delivers brevity, clarity, and comprehensiveness in translating medical image readings<br>- ChatGPT type system uses medical images to generate readings<br>- Will be a great help in analyzing options, guiding patients through their daily routines, and providing psychological counseling | - ChatGPT's translation lacks completeness and may miss key points<br>- ChatGPT's responses are inconsistent or uncertain, and information can be oversimplified or lost, even in the same prompt |
| 9 | - Make medical image readings more accessible and understandable in online patient portals, allowing patients to receive results directly without physician guidance | - Concerned about potential harm, overall quality, and sending reports to patients |
| 10 | - The latest LLMs, ChatGPT-3.5 and ChatGPT-4.0, provide a simpler and more efficient way to generate structured reports | - High satisfaction and accuracy on some tasks, but not perfect in terms of evaluation and unsatisfactory in terms of stability<br>- High rate of incorrect opinions in ChatGPT-3.5 vs. ChatGPT-4.0 in management opinion generation, and confusing manipulated opinions can mislead junior doctors |
| 11 | - ChatGPT achieves high accuracy in interpreting medical imaging readings and generating content<br>- Our study supports the above view and finds that ChatGPT can effectively interpret medical image readings and provide additional information relevant to the report | - The findings that ChatGPT-generated Hindi translations were inadequate for effective patient communication were due to several key factors<br>- ChatGPT lacks depth and breadth of knowledge and scores below average medical student level |
| 12 | - Demonstrated that the proposed method showed improved performance in predicting medical imaging findings | - Study identifies three main limitations of the model<br>- Unfamiliarity with the body due to limited training data. Poor performance on sites<br>- De-identified dataset limits follow-up advice from physicians<br>- Resource and time constraints: lack of GPUs limits training |
| 13 | - Suggests focusing on areas where imperfect AI with moderate performance can still create value | - Human verification and accountability issues<br>- Need near-perfect performance in clinical applications<br>- Need to design contextualized "workflow tools"<br>- Too broad for specific applications requires customization<br>- Demonstrate clear benefits of change or innovation<br>- Explanations of how AI works are still limited |
| 14 | - The potential of AI and LLM in medical imaging and diagnostics. Performance of domain-specific visual-language models in radiology outperformed ChatGPT. | - ChatGPT's data is not medical-specific due to limitations of internet-informed reinforcement learning<br>- Due to usage policies and ethical boundaries, ChatGPT is designed to deny requests for specialized interpretation of medical images.<br>- Hallucinations can cause serious problems in real-world clinical applications |

| Separation | Opportunities | Challenges |
|---|---|---|
| 15 | - GPT-4V demonstrates excellent ability to distinguish between different medical image modalities and anatomical structures<br>- In some tests, models showed a remarkable ability to analyze research findings and provide valuable insights and criteria that align with expert knowledge | - GPT-4V struggled to accurately diagnose diseases and generate comprehensive medical reports. In the worst cases, it appeared to "hallucinate" facts and make inference errors, generating responses that overlooked important information in the image input.<br>- GPT-4V features need to be further explored and refined through prompted engineering |

## Discussion

This study aimed to evaluate the clinical value of large-scale generative AI applications in radiology, with a focus on accuracy, workflow efficiency, and patient outcomes. The research spanned publications from January 2017 to April 2024, both in Korea and internationally. By examining 15 selected studies, this research dissected the general characteristics of the literature, the technical attributes and types of research subjects, the clinical value and outcomes of the subjects, and the opportunities and challenges they present.

First, the general characterization of the literature reveals that the majority of studies are based on radiology reports. Historically, medical image analysis research has been dominated by solutions employing CNN-based deep learning techniques to detect lesions in images. However, these models face clinical application limitations as they are effective only in detecting labeled lesions and do not encompass the entire reading process. The advent of Large Language Models (LLMs) has shifted AI research in radiology towards a focus on reading-centered tasks, such as summarizing, reorganizing, and translating radiology reports. Nevertheless, dealing solely with medical images or readings addresses only a portion of the radiologist's workload. The emergence of multimodal models that integrate these functionalities allows for an expanded role of AI beyond technical constraints.<Figure 2> Vision-language models can simultaneously handle medical images and readings, integrating all reading processes comprehensively. However, challenges such as lower sensitivity and specificity compared to CNN-based models, biases in training data, and difficulties in performance evaluation persist. Future research should prioritize advancing multimodal AI solutions to more complex modalities like CT and MRI, which demand substantial technological advancements and iterative development.

**Figure 2: Integrating the reading process of a multimodal model into the readout process**

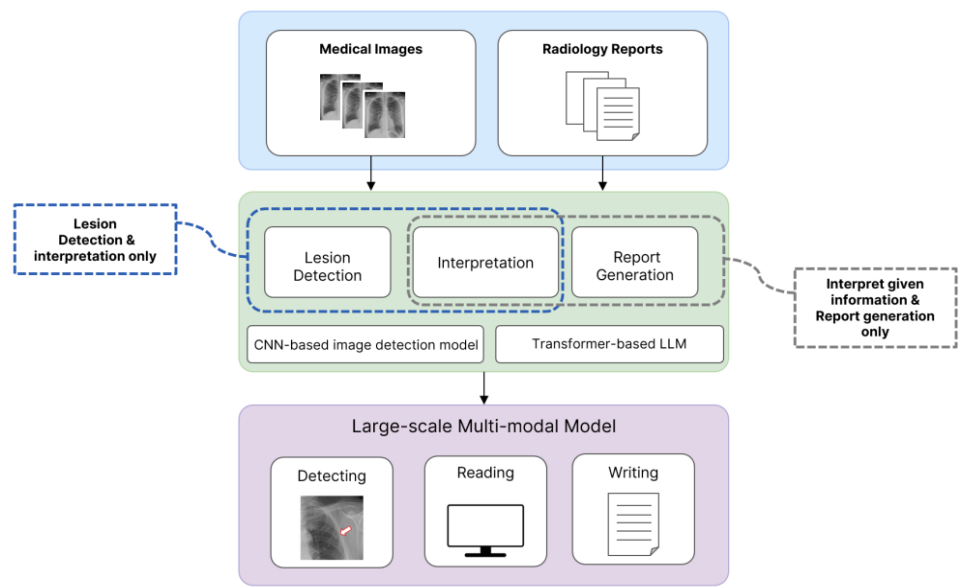

Second, the technical characteristics and types of research subjects are associated with the type of AI developer, modality, and AI model version. Most studies have utilized OpenAI's LLM, showing no particular bias in terms of modality type. The performance related to reading accuracy and error correction gradually increases as the GPT version improves, while functional advancements such as summarization and grammatical error correction do not differ.   However, it is important to note that OpenAI is not a healthcare-specific model because it aims for AGI. Currently, a variety of super-generative AI models are emerging, including some that are specific to the medical field, such as Google's Med-Gemini. While these models are the first of their kind, there is no clinical evidence that they outperform other ground truth models, or that they outperform human experts. However, we believe that it is significant from the perspective of aggregating the medical field by simultaneously learning from a large amount of medical data (medical images, EMR, vital signs, medical books, etc.) of various types. These AGIs use MoE (Mixture of Expert), a machine learning technique that combines multiple expert models to achieve better performance predictions. In the medical field, we predict the emergence of domain-specific foundation models and individual solutions based on or composed of domain-specific foundation models that specialize in each branch of medicine. The sum of these models will ultimately lead to the completion of AGIs in the form of <Figure 3>. Even in the medical imaging field, which has been the fastest to apply AI in the medical field, the emergence of a visual language-based model for chest X-ray, such as the case of Microsoft, is showing movement, and the emergence of a foundation model that integrates all modalities (Radiology domain specific foundation model) is expected soon. This move is expected to be followed by a sequential application of the model that is suitable for multi-modal applications, not only in the field of radiology.

**Figure 3: Medical AGI's Mixture of Expert(MoE)[36] hypothetical.**

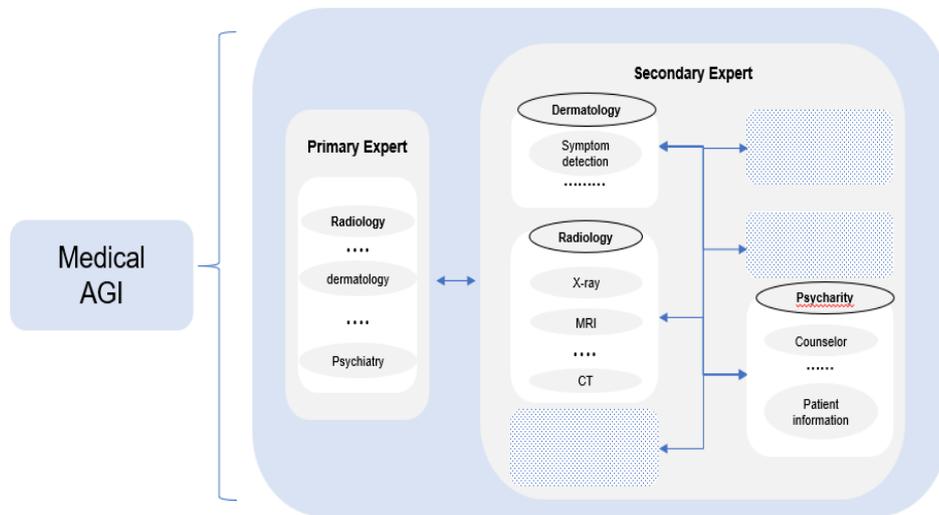

Third, when we look at the clinical value of the research, we find that of the three factors, Accuracy is the most common with 7 studies (54%), followed by better outcomes for Patient with 5 studies (38%), and Workflow efficiency with 4 studies (31%). The majority of LLM-related studies evaluated error correction in readings, conversion of readings into structured formats, and automatic generation of impressions. For certain parts of the medical image reading process, LLMs have been shown to perform well for time-consuming, tedious, or repetitive tasks that can be performed by human readers, and in terms of diagnostic performance, LLMs have not yet caught up with human performance. Therefore, LLMs are mainly used to fill in for humans in certain areas where natural language is utilized, or to augment human performance, i.e., to achieve certain results with a more efficient investment of time and money. On the other hand, research on multimodal models tends to focus on experiments that integrate the entire process (medical image analysis and reading generation), rather than just a specific part of the reading process, and involve human evaluation. This seems to indicate that the researchers' view of the role of AI is shifting from Assistance to Replacement for specific tasks. Although there were many opinions that the GPT-4V used in the study was generally insufficient in terms of performance evaluation alone, the performance of multimodal AI specialized in chest X-ray such as KARA-CXR was evaluated to be superior to that of GPT-4V, and based on the results of the study that the quality of reading (readability, grammatical errors, etc.) of these LLMs and multimodal models is similar to that of human experts, it can be said that multimodal generative AI specialized in medical imaging may replace radiologists in terms of function in the future.

In terms of patient usability, the literature focused on how well LLMs could translate the medical terms in the reading into a form that patients could understand. Most of the studies in the literature gave LLMs high marks for their translation capabilities and expected them to serve as a tool to increase patient-doctor interaction. In terms of technology, the recent release of the GPT-4o shows that the current state of the art of AI can convert text into various forms such as video, sound, and images, and the way and speed of interaction are greatly improving. Therefore, in the future, it may be possible to go beyond medical language translation and allow patients and A.I. to discuss readings in real time. However, there are still concerns about misleading information due to issues such as hallucinations, as the translation is not yet perfect. Even if this issue is resolved, there are some doctors who are reluctant to deliver medical information to patients, and there are cases where patients may misuse translated medical information by projecting their own intentions, so regulatory guidelines are needed.

Fourth, it's about the opportunities and challenges of the subject of the study.

First, in terms of opportunities, researchers believe that generative AI can be used in a variety of areas, such as streamlining radiologists' work, reducing workload and time, using it in trainee education, and interacting with patients. On the other hand, challenges ranged from legal issues such as hallucination and privacy to institutional issues such as review and accountability of AI results, data bias, and fundamental limitations of generative AI. In this regard, the aforementioned - if a multimodal based medical imaging domain-specific AI foundation model emerges and the preliminary issues in the clinical, technical, and institutional sense are resolved - is expected to enable a wide range of applications within various medical fields. For example, in Low Middle Income Country, where there are no radiologists and the illiteracy rate is high, it is expected that there will come a time when it will be possible to utilize a multimodal LLM in medical imaging to perform readings and receive the results immediately, and to translate the results into a language that the patient can understand through a smartphone, and to have a voice conversation with an artificial intelligence model to help the patient understand and cope with the diagnosis situation. New workflows could also emerge where clinicians other than radiologists can interact directly with AI readings to search for rare cases or answer clinical questions. However, I would like to point out that the diagnostic performance of AI should be close to my definition of near perfect AI performance as a prerequisite for human replacement, i.e., as shown in the literature, radiologists should be able to correct only about 1 in 10 AI-generated readings, which requires more quality data to improve performance and new clinical evaluation metrics - qualitative and quantitative metrics to measure reader satisfaction. Medical data, in particular, is not readily available in developed countries due to the closed behavior of healthcare organizations, data ownership issues, and difficult anonymization guidelines. In addition, high clinical perfection and high reader satisfaction are different issues, so personalization of the generated reading style and reinforcement learning through feedback (RLHF-Reinforcement learning of Human feedback) are essential to increase reader satisfaction and AI reading acceptability. In addition, in order to be applied to actual clinical sites, it should be possible to install it in the hospital in the form of an on-premise server that can solve information protection security issues, and it is essential to optimize and lighten the algorithm to realize it. In addition, economic issues such as the cost of high-end computing using GPUs and excessive power consumption (AI infrastructure) must also be addressed. From this perspective, it is expected that it will take a long time to resolve the three-dimensional issues of technological advancement, clinical perfection, data supply and demand, development and measurement of new evaluation metrics, personalization and security issues, and affordability. In the longer term, as algorithms mature in performance and clinical certainty is established, there will be a backlash from specialists against radiologist passing, where readings are sent directly to clinicians, bypassing the radiologist. This issue is a double-edged sword of solving the radiology shortage and improving algorithmic performance, so it falls in line with the old "AI threatens human jobs" argument. It is also worth noting that there is an inherent issue of patent infringement on the underlying AI technology, which is monopolized by the US big techs. In particular, Google invented self-attention, which is used for transformers in the GPT model, and allowed it to be widely used as open source, and various models were derived from it, leading to the development of generative AI today. As such, the AI industry is characterized by a culture of open source, where all research and inventions are shared and can be used by everyone without any restrictions. However, the underlying technology, the AI algorithms that can best implement them, and the vast infrastructure are in the hands of the U.S. Big Tech monopoly, and there is a risk that when the development of AI reaches a certain point, it will lead to so-called "closure and monopolization," which is the collapse of various open source communities due to non-disclosure of underlying technology and source code, potential patent attacks, etc. These issues are being addressed in the form of tacit agreements, but a transnational agreement that fosters these issues and provides institutional safeguards is a prerequisite.

Finally, there is a view that the fundamental problem with generative AI, namely hallucinations, may be accelerated and become harder to solve. In particular, in the context of medical AI, where accuracy is critical, hallucinations are directly related to the trustworthiness of the solution. Currently, as the number of parameters in the model increases, it becomes larger and larger, and the complexity of the algorithm increases, making it more difficult to understand how it works. While we've seen a number of recent moves to address these challenges, the industry consensus is that they are far from complete. Instead, various approaches are being used to mitigate the phenomenon, including improving data quality, improving models, incorporating user feedback, and implementing explainable AI. As AI diagnostic solutions are classified as software as medical devices (SaMD) under the control of regulatory agencies such as the Ministry of Food and Drug Safety, they are in the process of establishing standards and guidelines to reduce the problem of hallucinations through the process of evaluating and approving the reliability and safety of such applications. However, since the standards vary by country/region, and the US FDA, a representative of regulatory agencies, views AI medical devices with the most conservative nuances, many challenges are expected before they are used in actual clinical practice. There is also a move to solve the problem from a technical perspective. Recently, Anthropic, a new company founded by researchers from Open AI, announced that it understood the working principle of LLM, and it is hoped that the progress of this research will enter a new phase of the Explainable AI era and solve fundamental problems such as hallucinations.

# Method

In conducting this scoping review, We followed Arksey and O'Malley's definition[37], a scoping review[38] maps key concepts and evidence in a research area. A thematic review was chosen over a systematic review to systematically synthesize and clarify research amid the rise of AI paradigms like ChatGPT.

**Databases searched**

A systematic review was conducted using databases that reflect the latest trends and allow for advanced searching, including PubMed, EMbase, IEEE, and Google Scholar. The selection of databases and keywords was finalized with the consultation of a specialist in systematic reviews.

**Inclusion and exclusion criteria**

We utilized a variety of search terms encompassing concepts related to multi-modal generative AI, clinical value, and radiology to identify a broad range of peer-reviewed original records on these topics. The literature search included records published between January 1, 2017, and April 30, 2024. Records prior to 2023 were removed as duplicates and irrelevant because, although the Transformer architecture was first introduced in 2017, substantial related research began only after the introduction of ChatGPT in November 2022. Clinical value(for our context, Clinical value means 'diagnostic accuracy', 'Workflow efficiency' and 'better patient outcome') was defined based on the study by Johnbury (1991). We included only records describing original studies. Records without full text, commentaries, background information, articles not evaluated by clinicians (not related clinical value), and research protocols related to AI algorithms were excluded. Additionally, our search primarily included records that provided quantitative evaluations of generative AI applications in radiology.

**Study screening and selection**

Records identified by the above database searches were entered into Microsoft Excel 2019 for further title and abstract review. Inclusion and exclusion criteria were identified following the PCC (Population, Concept, Context) guidelines [39] Two reviewers (InWoo Seo, EunKyung Bae) independently screened the titles and abstracts to select articles fully meeting the inclusion criteria related to the application of multi-modal generative AI, clinical value, and radiology. Once relevant articles were identified, the reviewers (InWoo Seo, EunKyung Bae) screened all full texts to exclude those articles that did not meet the inclusion criteria based on the full-text review.

**DATA AVAILABILITY**

All data generated and analyzed during this study are included in the article and its
supplementary information files.